\documentclass[11pt,a4paper]{article}
\usepackage{times,latexsym}
\usepackage{url}
\usepackage{times}
\usepackage{soul}
\usepackage{url}
\usepackage[hidelinks]{hyperref}
\usepackage[utf8]{inputenc}
\usepackage[small]{caption}
\usepackage{graphicx}
\usepackage{amsmath}
\usepackage{amsthm}
\usepackage{booktabs}
\usepackage{algorithm}
\usepackage{algorithmic}
\usepackage{multirow}
\usepackage{pifont}
\newcommand{\xmark}{\ding{55}}%
\newcommand{\cmark}{\ding{51}}%
\usepackage[T1]{fontenc}
\newcommand{\linestack}[1]{\def\arraystretch{0.7}\begin{tabular}[c]{@{}c@{}} #1 \end{tabular}}
\usepackage[acceptedWithA]{tacl2018v2}

\usepackage{xspace,mfirstuc,tabulary}

\newif\iftaclinstructions
\taclinstructionsfalse 
\iftaclinstructions
\renewcommand{\confidential}{}
\renewcommand{\anonsubtext}{(No author info supplied here, for consistency with
TACL-submission anonymization requirements)}
\newcommand{\instr}
\fi

\iftaclpubformat 

\else

\fi

\title{Advances in Multi-turn Dialogue Comprehension: A Survey}

\author{
 Zhuosheng Zhang and
Hai Zhao\\
$^1$Department of Computer Science and Engineering, Shanghai Jiao Tong University\\
$^2$Key Laboratory of Shanghai Education Commission for Intelligent Interaction\\
	and Cognitive Engineering, Shanghai Jiao Tong University, Shanghai, China\\
$^3$MoE Key Lab of Artificial Intelligence, AI Institute, Shanghai Jiao Tong University, Shanghai, China\\
  {\sf zhangzs@sjtu.edu.cn, zhaohai@cs.sjtu.edu.cn} \\
}

\date{}

\begin{document}
\maketitle
\begin{abstract}
Training machines to understand natural language and interact with humans is an elusive and essential task of artificial intelligence. A diversity of dialogue systems has been designed with the rapid development of deep learning techniques, especially the recent pre-trained language models (PrLMs). Among these studies, the fundamental yet challenging {{type of task}} is dialogue comprehension whose role is to teach the machines to read and comprehend the dialogue context before responding. In this paper, we review the previous methods from the {{technical}} perspective of dialogue modeling for the dialogue comprehension {{task}}. We summarize the characteristics and challenges of dialogue comprehension in contrast to plain-text reading comprehension. Then, we discuss three typical patterns of dialogue modeling. In addition, we categorize dialogue-related pre-training techniques which are employed to enhance PrLMs in dialogue scenarios. Finally, we highlight the technical advances in recent years and point out the lessons from the empirical analysis and the prospects towards a new frontier of researches.
\end{abstract}

\section{Introduction}
{{Building an intelligent dialogue system that can naturally and meaningfully communicate with humans is a long-standing goal of artificial intelligence (AI) and has been drawing increasing interest from both academia and industry areas due to its potential impact and alluring commercial values. It is a classic topic of human-machine interaction that has a long history. Before computer science and artificial intelligence were categorized into various specific branches, dialogue has become a critical research topic with clear application scenario, as a phenomenon. Dialogue also serves as the important applications area for pragmatics \cite{leech2003pragmatics} and the Turing Test \cite{turing1950computing}.}}

\begin{figure*}
\centering
\includegraphics[width=1.0\textwidth]{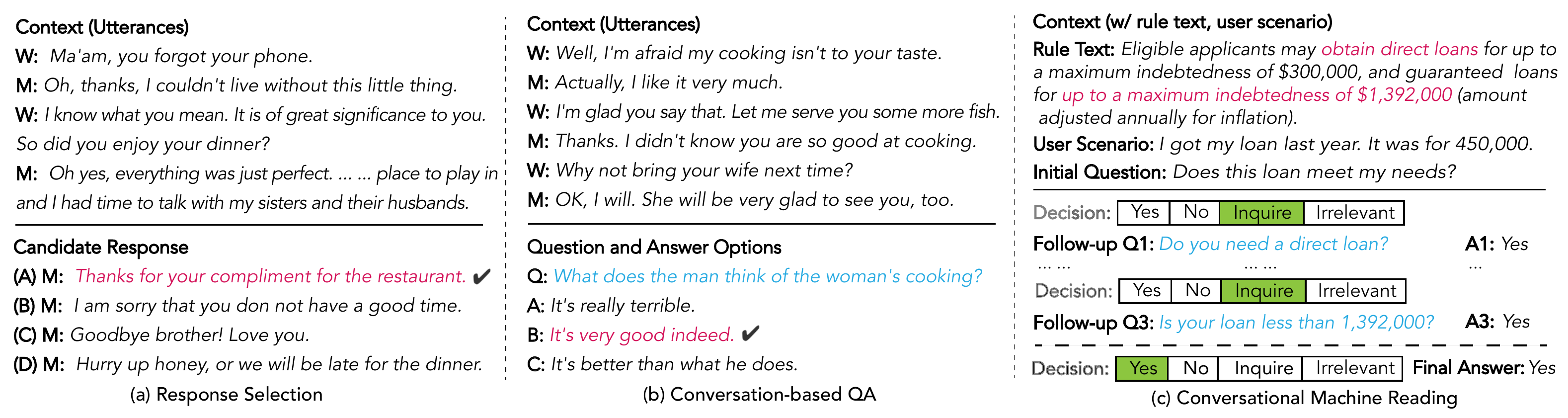}
\caption{Examples for various dialogue comprehension tasks including response selection, conversation-based QA, and conversational machine reading (w/ decision making and question generation).}
\label{dialogue_exp}
\end{figure*}

{{Traditional methods were proposed to help users complete specific tasks with pre-defined hand-crafted templates after analyzing the scenario of the input utterances \cite{weizenbaum1966eliza,colby1971artificial}, which are recognized as rule-based methods \cite{chizhik2020challenges}. However, the growth in this area is hindered by the problem of data scarcity as these systems are expected to learn linguistic knowledge, decision making, and question answering from insufficient amounts of high-quality corpora \cite{zaib2020short}. To alleviate the scarcity, }} a variety of tasks have been proposed such as response selection \cite{Loweubuntu,Wudouban,Zhangedc}, conversation-based question answering (QA) \cite{sun2019dream,reddy2019coqa,choi2018quac}, decision making and question generation \cite{saeidi-etal-2018-interpretation}. Examples are shown in Figure \ref{dialogue_exp}. 

\begin{figure*}
\centering
\includegraphics[width=1.0\textwidth]{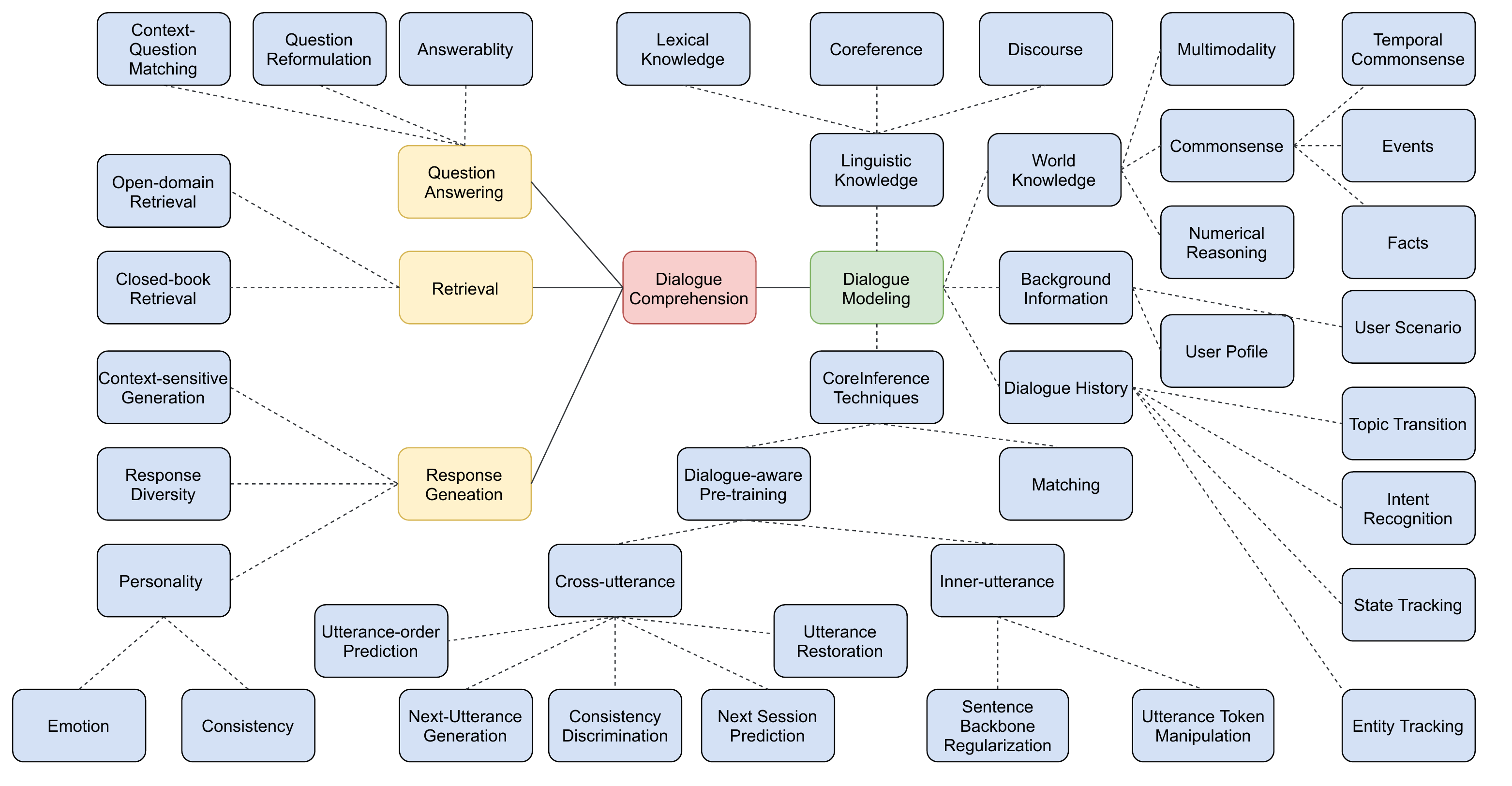}
\caption{{{Overview of dialogue comprehension as phenomenon. The left part illustrates the task-related skills for dialogue comprehension. The right part presents the typical techniques in view of dialogue modeling.}}}
\label{fig:dlg_phen}
\end{figure*}

 Recently, with the development of deep learning methods \cite{serban2016building}, especially the recent pre-trained language models (PrLMs) \cite{devlin2019bert,YinhanLiuroberta,yang2019xlnet,lan2019albert,clark2019electra}, {{the capacity of neural models has been boosted dramatically. {{However, most studies focus on an individual task like response retrieval or generation. }}Stimulated by the interest towards building more generally effective and comprehensive systems to solve real-world problems, }} traditional natural language processing (NLP) tasks, including the dialogue-related tasks, have been undergoing a fast transformation, where those tasks tend to be crossed and unified in form \cite{zhang2020mrc}. {{Therefore, we may view the major dialogue tasks in a general format of dialogue comprehension: given contexts, a system is required to understand the contexts, and then reply or answer questions. The reply can be derived from retrieval or generation. As a generic concept of measuring the ability to understand dialogues, as opposed to static texts, dialogue comprehension has broad inspirations to the NLP/AI community as shown in Figure \ref{fig:dlg_phen}.}}


{{Among the dialogue comprehension studies, the basic technique is dialogue modeling which focuses on how to encode the dialogue context effectively and efficiently to solve the tasks, thus we regard dialogue modeling as the technical aspect of dialogue comprehension.}} Early techniques mainly focus on the matching mechanisms between the pairwise sequence of dialogue context and candidate response or question \cite{Wudouban,Zhangedc,huang2018flowqa}. {{Recently, PrLMs have shown impressive evaluation results for various downstream NLP tasks \cite{devlin2019bert} including dialogue comprehension. They handle the whole texts as a linear sequence of successive tokens and capture the contextualized representations of those tokens through self-attention \cite{qu2019bert,liu2020hisbert,Gusabert,xu2021learning}. The word embeddings derived by these language models are pre-trained on large corpora. Providing fine-grained contextual embedding, these pre-trained models could be either easily applied to downstream models as the encoder or used for fine-tuning. Besides employing PrLMs for fine-tuning, there also emerges interest in designing dialogue-motivated self-supervised tasks for pre-training.
}}

In this survey, we review the previous studies of dialogue comprehension in the perspective of modeling the dialogue tasks as a two-stage Encoder-Decoder framework inspired by the advance of PrLMs and machine reading comprehension \cite{zhang2020mrc}, in which way we bridge the gap between the dialogue modeling and comprehension, and hopefully benefit the future researches with the cutting-edge PrLMs. In detail, we will discuss both sides of architecture designs and the pre-training strategies. We summarize the technical advances in recent years and highlight the lessons we can learn from the empirical analysis and the prospects towards a new frontier of researches. Compared with existing surveys that focus on specific dialogue tasks \cite{zaib2020short,huang2020challenges,fan2020survey,qin2021survey}, this work is task-agnostic that discusses the common patterns and trends of dialogue systems in the scope of dialogue comprehension, in order to bridge the gap between different tasks so that those research lines can learn the highlights from each other.

\section{Characteristics of Dialogues}\label{sec:character}
In contrast to plain-text reading comprehension like SQuAD \cite{Rajpurkar2016SQuAD}, {{a multi-turn conversation is intuitively associated with spoken (as opposed to written) language and also is also interactive that involves multiple speakers, intentions, topics, thus the utterances are full of transitions.}}

1) {{\textbf{Speaker interaction}.}} The transition of speakers in conversations is in a random order, breaking the continuity as that in common non-dialogue texts due to the presence of crossing dependencies which are commonplace in a multi-party chat. 

2) {{\textbf{Topic Transition}.}} There may be multiple dialogue topics happening simultaneously within one dialogue history and topic drift is common and hard to detect in spoken conversations. Therefore, the multi-party dialogue appears discourse dependency relations between non-adjacent utterances, which leads up to a complex discourse structure.

3) {{\textbf{Colloquialism}.}} Dialogue is colloquial, and it takes fewer efforts to speak than to write, resulting in the dialogue context rich in component ellipsis and information redundancy. However, during a conversation, the speakers cannot retract what has been said, which easily leads to self-contradiction, requiring more context, especially clarifications to fully understand the dialogue.

4) {{\textbf{Timeliness}.}} The importance of each utterance towards the expected reply is different, which makes the utterances contribute to the final response in dramatic diversity. Therefore, the order of utterance influences the dialogue modeling. In general, the latest utterances would be more critical \cite{Zhangedc,zhang2021kkt}.

\begin{figure*}[t]
\centering
\includegraphics[width=0.98\textwidth]{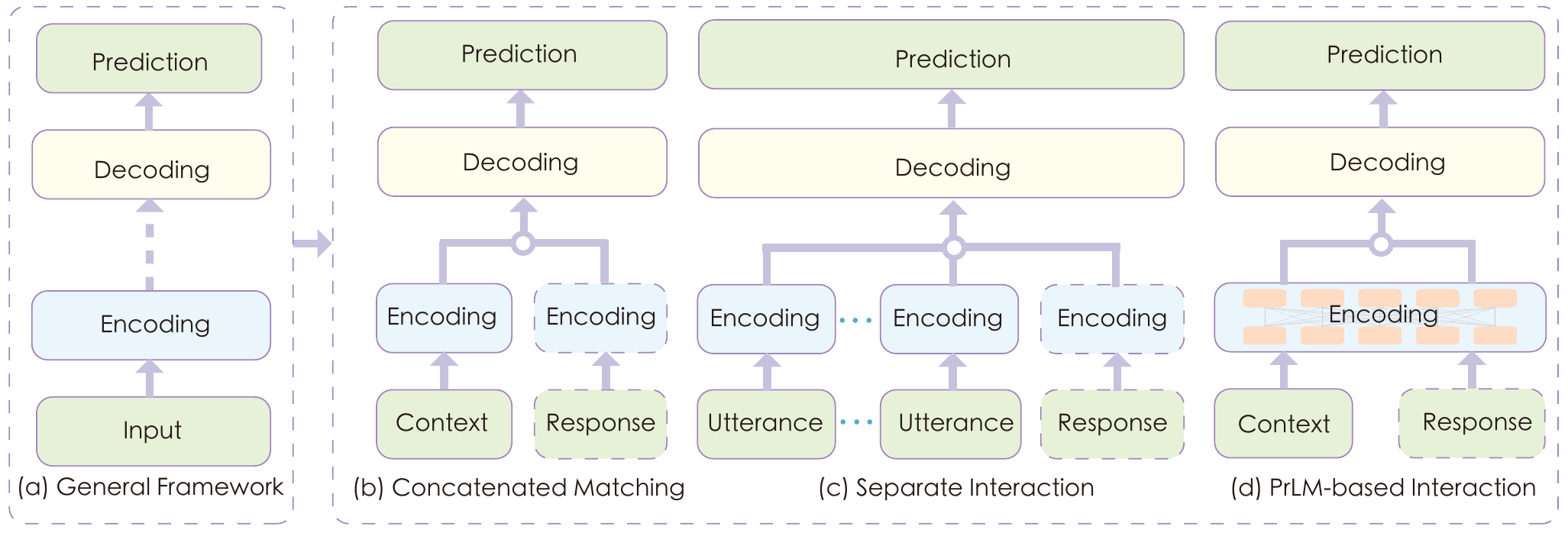}
\caption{{{Dialogue modeling framework. The dispensable parts are marked in dashed lines. Without the response as input, the framework is then applicable to dialogue generation tasks.}}}
\label{dialogue_fm}
\end{figure*}

\begin{figure*}
\centering
\includegraphics[width=0.98\textwidth]{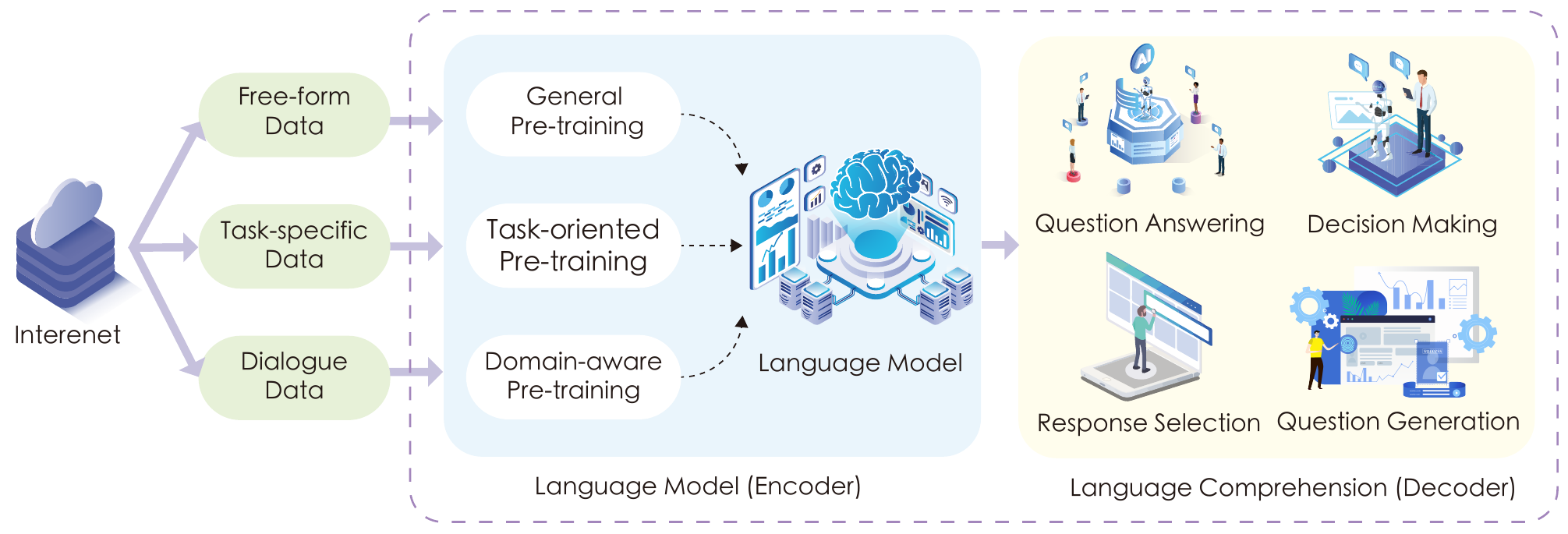}
\caption{{{Dialogue-related pre-training. There are three kinds of dialogue-related language modeling strategies, namely, general-purpose pre-training, domain-aware pre-training, and task-oriented pre-training.}}}
\label{dialogue_lm}
\end{figure*}

\section{Methodology}
\subsection{Problem Formulation}
{{Although existing studies of dialogue comprehension commonly design independent systems for each downstream tasks, we find that dialogue systems can be generally formulated as an Encoder-Decoder framework where an encoder is employed to understand the dialogue context and the decoder is for giving the response.}} The backbone of the encoder can be either a recurrent neural network such as LSTM \cite{hochreiter1997long}, or a pre-trained language model such as BERT \cite{devlin2019bert}. The decoder can be simple as a dense layer for discriminative tasks such as candidate response classification, or part of the seq2seq architecture \cite{bahdanau2014neural} for question or response generation. {{We could reach the view that dialogue comprehension tasks, especially dialogue generation, share essential similarity with machine translation \cite{sennrich2016neural}, from which such unified modeling view can be used to help develop better translation and dialogue generation models from either side of advances.}}

Here we take two typical dialogue comprehension tasks, i.e., response selection \cite{Loweubuntu,Wudouban,Zhangedc,mutual} and conversation-based QA \cite{sun2019dream,reddy2019coqa,choi2018quac}, as examples to show the general technical patterns to gain insights, which would also hopefully facilitate other dialogue-related tasks.\footnote{Actually, there are other dialogue comprehension tasks such as decision making and question generation that share similar formations as the fundamental part is still the dialogue context modeling. We only elaborate on the two examples to save space.}

Suppose that we have a dataset ${D} = \{( \textup{C}_i, \textup{X}_i; \textup{Y}_i)\}^N_{i=1}$, where ${\textup{C}_i} = \{u_{i,1},...,u_{i,n_i}\}$ represents the dialogue context with $\{u_{i,k}\}^{n_i}_{k=1}$ as utterances. $\textup{X}_i$ is a task-specific paired input, which can be either the candidate response ${R}$ for response selection, or the question $Q$ for conversation-based QA. $\textup{Y}_i$ denotes the model prediction.

\textit{Response Selection} involves the pairwise input with ${R}$ as a candidate response. The goal is to learn a discriminator $g(\cdot,\cdot)$ from ${D}$, and at the inference phase, given the context ${C}$ and response ${R}$, we use the discriminator to calculate $Y = g({C},{R})$ as their matching score. 

\textit{Conversation-based QA} aims to answer questions given the dialogue context. Let ${Q}$ denotes the question ${Q}$. The goal is to learn a discriminator $g(C, Q)$ from ${D}$ to extract the answer span from the context or select the right option from a candidate answer set. 

In the input encoding perspective, since both of the tasks share the paired inputs of either $\{{C};{R}\}$ or $\{{C};{Q}\}$, we simplify the formulation by focusing the response selection task, i.e., replacing ${R}$ with ${Q}$ can directly transform into the QA task and {{without the response as input, the framework is then applicable to dialogue generation tasks.}}

\subsection{Dialogue Modeling Framework}
As shown in Figure \ref{dialogue_fm}, {{we review the previous studies of dialogue comprehension in the perspective of modeling the dialogue tasks as a two-stage Encoder-Decoder framework.}}
The methods of dialogue modeling can be categorized into three patterns: 1) concatenated matching; 2) separate interaction; 3) PrLM-based interaction.  

\paragraph{{{Framework 1:}} Concatenated Matching} The early methods \cite{kadlec2015improved} treated the dialogue context as a whole by concatenating all previous utterances and last utterance as the context representation and then computed the matching degree score based on the context representation to encode candidate response \cite{Loweubuntu}:
\begin{equation}
\begin{split}
    &\textup{EC} = \textup{Encoder}({C});\\
    &\textup{ER} = \textup{Encoder}({R});\\
    &\textup{Y} = \textup{Decoder}(\textup{EC};\textup{ER});\\
\end{split}
\end{equation}
where \textup{Encoder} is used to encode the raw texts into contextualized representations. \textup{Decoder} is the module that transforms the contextualized representations to model predictions (Y), which depends on the tasks. For response selection, it can be the attention-based module that calculate the matching score between EC and ER.


\paragraph{{{Framework 2:}} Separate Interaction} With the bloom of attention-based pairwise matching mechanisms, researchers soon find it effective by calculating different levels of interactions between the dialogue context and response. The major research topic is how to improve the semantic matching between the dialogue context and candidate response. For example, \citeauthor{zhou2016multi} \shortcite{zhou2016multi} performed context-response matching with a multi-view model on both word-level and utterance level. \citeauthor{Wudouban} \shortcite{Wudouban} proposed to capture utterances relationship and contextual information by matching a response with each utterance in the context. Those methods can be unified by the view similar to the above concatenated matching:
\begin{equation}
\begin{split}
    &\textup{EU}_i = \textup{Encoder}(u_i);\\
    &\textup{ER} = \textup{Encoder}({R});\\
    &\textup{I} = \textup{ATT}([\textup{EU}_1,\dots,\textup{EU}_n];\textup{ER});\\
    &\textup{Y} = \textup{Decoder}(\textup{I});\\
\end{split}
\end{equation}
where \textup{ATT} denotes the attention-based interactions, which can be pairwise attention, self attention, or the combinations.

\paragraph{{{Framework 3:}} PrLM-based Interaction} PrLMs handle the whole input text as a linear sequence of successive tokens and implicitly capture the contextualized representations of those tokens through self-attention \cite{devlin2019bert}. Given the context ${C}$ and response ${R}$, we concatenate all utterances in the context and the response candidate as a single, consecutive token sequence with special tokens separating them, and then encode the text sequence by a PrLM:
\begin{equation}
\begin{split}
    &\textup{EC} = \textup{Encoder}(\texttt{[CLS]}{C}  \texttt{[SEP]} {R} \texttt{[SEP]});\\
    &\textup{Y} = \textup{Decoder}(\textup{EC});\\
\end{split}
\end{equation}
where \texttt{[CLS]} and \texttt{[SEP]} are special tokens. 

{{\paragraph{Comparison of Three Frameworks}
In the early stages of studies that lack computational sources, concatenated matching has the advantage of efficiency, which encodes the context as a whole with a simple structure and directly feeds it to the decoder. With the rapid spread of attention mechanisms, separate interaction has become mainstream and is generally better than concatenated matching because the relationships between utterances and between utterances and response can be sufficiently captured after the fine-grained attention-based interaction. PrLM-based models further extend the advantage of interaction by conducting multi-layer word-by-word interaction over the context and the response. With another benefit from pre-training on large-scale corpora through self-supervised tasks, PrLM-based models significantly outperform the conventional two frameworks. However, the latter two interaction-based methods would be less efficient for real-world applications due to the cost of the heavy computation. Fortunately, it is possible to keep the effectiveness and efficiency at the same time. Inspired by the recent studies of dense retrieval \cite{seo2019real,karpukhin2020dense,zhang2021oscar} and the fact that dialogue histories are often used repeatedly, a potential solution is to pre-compute their representations for latter indexing, which allows for fast real-time inference in a production setup, giving an improved trade-off between accuracy and speed \cite{humeau2019poly}.}}

\begin{table*}[t]
\centering
\setlength{\tabcolsep}{2pt}
\footnotesize
  {\begin{tabular}{l c c c c c c}
    \toprule
    \multirow{2}{*}{\textbf{Model}} &
    \multicolumn{2}{c}{\textbf{Standard Language Modeling}}
 & \multicolumn{2}{c}{\textbf{Task-oriented Objective}} & \multicolumn{2}{c}{\textbf{Application Task}}\\
     & Masked LM & $n$-gram LM  & Cross-utterance & Inner-utterance & Retrieval & Generation \\
    \midrule
    BERT-VFT \cite{whang2020domain} & \cmark & & \cmark &  & \cmark &  \\
    DIALOGPT \cite{DialoGPTLG2020Zhang} &  & \cmark & & & & \cmark \\
    SA-BERT \cite{Gusabert} & \cmark & & \cmark & & \cmark &\\
    PoDS \cite{zhang2021kkt} & \cmark & & \cmark & & \cmark & \\
    DAPO \cite{Lidapo} & \cmark & & \cmark & & \cmark & \\
    DCM \cite{li2021deep} & \cmark & & \cmark & & \cmark &\\
    \citet{zhao2020learning} & & \cmark & \cmark & \cmark & & \cmark\\
    SPIDER \cite{zhang-zhao-2021-structural}& \cmark & & \cmark & \cmark & \cmark & \\
    UMS \cite{whang2021response} & \cmark & & \cmark & & \cmark & \\
    BERT-SL \cite{xu2021learning}& \cmark& & \cmark & & \cmark &\\
  \bottomrule
  \end{tabular}}
  \caption{{{Dialogue-related pre-training methods. \textit{Application task} shows the evaluated task as reported in the corresponding literature. Note that some methods are applicable for both tasks though only evaluated on one of them.}}
  } 
  \label{table:topm_results}
\end{table*}

\subsection{Dialogue-related Pre-training}
Although the PrLMs demonstrate superior performance due to their strong representation ability from self-supervised pre-training, it is still challenging to effectively capture task-related knowledge during the detailed task-specific training \cite{gururangan-etal-2020-dont}. Generally, directly using PrLMs would be suboptimal to model dialogue tasks which holds exclusive text features that plain text for PrLM training may hardly embody. Besides, pre-training on general corpora has critical limitations if task datasets are highly domain-specific \cite{AnED2019Whang}, which cannot be sufficiently and accurately covered by the learned universal language representation.


Therefore, some researchers have tried to further pre-train PrLMs with general language modeling (LM) objectives on in-domain dialogue texts. The notable examples are BioBERT \cite{BioBERTAP2020Lee}, SciBERT \cite{SciBERTAP2019Beltagy}, Clinical-BERT \cite{ClinicalBERTMC2019Huang}, and DialoGPT \cite{DialoGPTLG2020Zhang}. As our work emphasizes dialogue comprehension tasks, which are more complex than other forms of texts like sentence pairs or essays, the corresponding training objective should be very carefully designed to fit the important elements of dialogues. As shown in Figure \ref{dialogue_lm}, there are three kinds of dialogue-related language modeling strategies, namely, \textit{general-purpose pre-training}, \textit{domain-aware pre-training}, and \textit{task-oriented pre-training}, among which self-supervised methods do not require additional annotation and can be easily applied into existing approaches.\footnote{Though general-purpose pre-training is not our major focus, we describe it here as the basic knowledge for completeness.} Typical examples are compared in Table \ref{table:topm_results}.

\paragraph{General-purpose Pre-training} 
As the standard pre-training procedure, PrLMs are pre-trained on large-scale domain-free texts and then used for fine-tuning according to the specific task needs. There are token-level and sentence-level objectives used in the general-purpose pre-training. BERT \cite{devlin2019bert} adopts Masked Language Modeling (MLM) as its pre-training objective. 
It first masks out some tokens from the input sentences and then trains the model to predict them by the rest of the tokens. 
There are derivatives of MLM like Permuted Language Modeling (PLM) in XLNet \cite{yang2019xlnet} and Sequence-to-Sequence MLM (Seq2Seq MLM) in MASS \cite{MASSMS2019Song} and T5 \cite{ExploringTL2019Raffel}. 
Next Sentence Prediction (NSP) is another widely used pre-training objective, which trains the model to distinguish whether two input sentences are continuous segments from the training corpus. Sentence Order Prediction (SOP) is one of the replacements of NSP. It requires models to tell whether two consecutive sentences are swapped or not and is first used in ALBERT \cite{lan2019albert}. 

\paragraph{Domain-aware Pre-training}
The PrLMs are pre-trained on a large text corpus to learn general language representations. To incorporate specific in-domain knowledge, adaptation on in-domain corpora, also known as domain-aware pre-training, is designed, which directly post-trains the original PrLMs using the dialogue-domain corpus \cite{whang2020effective,wu2020tod}. The most widely-used PrLM for domain-adaption in the dialogue field is BERT \cite{devlin2019bert}, whose pre-training is based on MLM and NSP loss functions. Although NSP has been shown trivial in RoBERTa \cite{YinhanLiuroberta} during general-purpose pre-training, it yields surprising gains in dialogue scenarios \cite{Lidapo}. The most plausible reason is that dialogue emphasizes the relevance between dialogue context and the subsequent response, which shares a similar goal with NSP. {{Notably, there are Seq2Seq (also known as Text2Text) Transformers pre-trained on massive conversational datasets that serve as the backbone of  conversation systems. Though those methods have the advantage of fluency, however, there are criticisms that they often suffer from factual incorrectness and hallucination of knowledge \cite{roller2021recipes}. A potential solution would be retrieving relevant knowledge and conditioning on dialogue turns \cite{shuster2021retrieval}. }}

\paragraph{Task-oriented Pre-training}
In contrast to the plain-text modeling as the focus of the PrLMs, dialogue texts involve multiple speakers and reflect special characteristics such as topic transitions and structural utterance dependencies as discussed in Section \ref{sec:character}. {{Inspired by such characteristics to imitate the real-world dialogues, recent studies are pondering the dialogue-specific self-supervised training objectives to model dialogue-related features. There are two categories of studies from the sides of cross-utterance and inner-utterance.
}}

1) {{\textbf{Cross-utterance}}}. Prior works have indicated that the order information would be important in the text representation, and the well-known {{NSP and SOP}} can be viewed as special cases of order prediction. Especially in the dialogue scenario, predicting the word order of utterance, as well as the utterance order in the context, has shown effectiveness in the dialogue modeling task \cite{kumar2020deep,gu2020dialogbert}, where the utterance order information is well restored from shuffled dialogue context. \citeauthor{li2021deep} \shortcite{li2021deep} designed a variant of NSP called next utterance prediction as a pre-training scheme to adapt BERT to accommodate the inherent context continuity underlying the multi-turn dialogue.  \citeauthor{whang2021response} \shortcite{whang2021response} proposed various utterance manipulation strategies including utterance insertion, deletion, and search to maintain dialog coherence. \citeauthor{xu2021learning} {{Similarly, \shortcite{xu2021learning} introduced four self-supervised tasks to explicitly model the cross-utterance relationships to improve coherence and consistency between the utterances, including next session prediction, utterance restoration, incoherence detection, and consistency discrimination.}}

2) {{\textbf{Inner-utterance
}. The other type of objectives is proposed as inner-utterance modeling, which has not attracted much attention. The intuition is to model the fact and events inside an utterance. \citet{zhang-zhao-2021-structural} introduced a sentence backbone regularization task as regularization to improve the factual correctness of summarized subject-verb-object triplets. \citet{zhao2020learning} proposed word order recovery and masked word recovery to enhance understanding of the sequential dependency among words and encourage more attention to the semantics of words to find better representations of words.}}

\begin{table*}\footnotesize
  \setlength{\tabcolsep}{2.6pt}
  \begin{tabular}{lcccccccccccc} \toprule
Name & Response Form & Size & Domain & Manually & \linestack{Task} \\ \midrule
Ubuntu \cite{Loweubuntu} & Choice & 2M & Technique & \xmark & Next Utterances Prediction \\
Douban \cite{Wudouban} & Choice & 1.1M & Open & \xmark & Next Utterances Prediction \\
ECD \cite{Zhangedc} & Choice & 1.02M  & E-commerce & \xmark & Next Utterances Prediction \\
PERSONA-CHAT \cite{zhang2018personalizing} & Choice & 164K & Persona& \cmark & Next Utterances Prediction \\
DailyDialog \cite{li2017dailydialog} & Free-form & 13K & Open & \cmark & Next Utterances Prediction \\
Dialogue NLI \cite{welleck2019dialogue} & Choice & 310K & Persona& \xmark & Next Utterances Prediction \\
MuTual \cite{mutual} & Choice & 8.8K & Open & \cmark & Next Utterances Prediction \\
DREAM \cite{sun2019dream} & Choice & 10K & Open & \cmark & Conversation-based QA \\
FriendsQA \cite{yang-choi-2019-friendsqa} & Span Extraction & 100K & Open & \cmark & Conversation-based QA \\
Molweni \cite{li2020molweni} &  \linestack{Span Extraction \\(+ unanswerable)} & 30K & Technique & \cmark & Conversation-based QA \\
QuAC \cite{choi2018quac} & Free-form (+ Yes/No) & 100K & Wikipedia & \cmark & Conversation-based QA \\ 
CoQA \cite{reddy2019coqa} & Free-form (+ Yes/No) & 127K & Wikipedia & \cmark & Conversation-based QA \\
ShARC \cite{saeidi-etal-2018-interpretation} & Yes/No$^*$ & 32K & web  snippet & \cmark & \linestack{Decision Making \& \\ Question Generation }  \\ 
\bottomrule
\end{tabular}
 \caption{{{Widely-used datasets for dialogue comprehension tasks.
 \textit{Response Form} shows the way to provide the response according to the official evaluation metrics in the corresponding literature. \textit{Size} indicates the size of the whole dataset, including training, development, and test sets. \textit{Manually} indicates that human writing of the
question or answers is involved in the data annotation process. Since dialogue corpus can often be used for both response selection and generation tasks, we use the term "next utterance prediction". }}
\label{sec:dataset_comp}
}
\end{table*}

\section{{{Dialogue Comprehension with Explicit Knowledge}}}
{{Dialogue contexts are colloquial and full of incomplete information, which requires a machine to refine the key information while reflecting the relevant details. Without background knowledge as a reference, the machine may merely capture limited information from the surface text of the dialogue context or query. Such knowledge is not limited to topics, emotions, and multimodal grounding. These sources can provide extra information beyond the textual dialogue context to enhance dialogue comprehension.}}

\subsection{Auxiliary Knowledge Grounding}
{{Linguistic knowledge has been shown important for dialogue modeling, such as syntax \cite{wang2015syntax,eshghi-etal-2017-bootstrapping} and discourse information \cite{galley2003discourse,ouyang2020dialogue}.}}
In addition, Various kinds of background knowledge can be adaptively grounded in dialogue modeling according to task requirements, including commonsense items  \cite{zhang2021kkt}, speaker relationships \cite{liu2020graph},  domain knowledge \cite{li2020alimekg}, from knowledge graphs to strengthen reasoning ability, persona-based attributes such as speaker identity, dialogue topic, speaker sentiments, to enrich the dialogue context \cite{olabiyi2019adversarial}, scenario information to provide the dialogue background \cite{ouyang2020dialogue}, etc. 

\subsection{Emotional Promotion}
Emotional feeling or sentiment is a critical characteristic to distinguish people from machines \cite{hsu2018emotionlines}. People's emotions are complex, and there are various complex emotional features such as metaphor and irony in the dialogue. Besides, the same expressions may have different meanings in different situations. Not only should the dialogue systems capture the user intents, topics transitions, and dialogue structures, but also they should be able to perceive the sentiment changes and even adjust the tone and guide the conversation according to the user emotional states, to provide a more friendly, acceptable, and empathetic conversations, which would be especially useful for building social bots and automatic e-commerce marketing. 

\subsection{Multilingual and Multimodal Dialogue}
As a natural interface of human and machine interaction, a dialogue system would be beneficial for people in different language backgrounds to communicate with each other. Besides the natural language texts, visual and audio sources are also effective carriers are can be incorporated with texts for comprehensive and immersed conversations. With the rapid development of multilingual and multimodal researches \cite{Qin2020CoSDA,firdaus2021aspect}, building an intelligent dialogue system is not elusive in the future.

\begin{table*}[t]
\centering
\setlength{\tabcolsep}{2pt}
\footnotesize
  {\begin{tabular}{lc cc c cc c c c c c}
    \toprule
    \multirow{2}{*}{\textbf{Model}} & \multirow{2}{*}{\textbf{Encoder}} & \multirow{2}{*}{\textbf{Augmentation}} & \multicolumn{3}{c}{\textbf{Ubuntu}} & \multicolumn{3}{c}{\textbf{Douban}} & \multicolumn{3}{c}{\textbf{E-commerce}}\\
     & & & $R@1$ & $R@2$ & $R@5$ & $R@1$ & $R@2$ & $R@5$ & $R@1$ & $R@2$ & $R@5$\\
    \midrule
    \multicolumn{12}{l}{\textit{Single-turn models with concatenated matching}}\\
    CNN \cite{kadlec2015improved} & CNN & - & 0.549 & 0.684 & 0.896 & 0.121 & 0.252 & 0.647 & 0.328 & 0.515 & 0.792\\
    LSTM \cite{kadlec2015improved}  & RNN & - & 0.638 & 0.784 & 0.949 & 0.187 & 0.343 & 0.720 & 0.365 & 0.536 & 0.828\\
    BiLSTM \cite{kadlec2015improved} & RNN & - & 0.630 & 0.780 & 0.944 &  0.184 & 0.330 & 0.716 & 0.365 & 0.536 & 0.825\\
    MV-LSTM \cite{wan2016match} & RNN & - & 0.653 & 0.804 & 0.946 &  0.202 & 0.351 & 0.710 & 0.412 & 0.591 & 0.857\\ 
    Match-LSTM \cite{wang2016learning}  & RNN & - & 0.653 & 0.799 & 0.944 &  0.202 & 0.348 & 0.720 & 0.410 & 0.590 & 0.858\\ 
    \midrule
    \multicolumn{12}{l}{\textit{Multi-turn matching network with separate interaction}}\\
    Multi-View \cite{zhou2016multi} & RNN & - & 0.662 & 0.801 & 0.951 &  0.202 & 0.350 & 0.729 & 0.421 & 0.601 & 0.861\\
    DL2R \cite{yan2016dl2r}  & RNN & - & 0.626 & 0.783 & 0.944 &  0.193 & 0.342 & 0.705 & 0.399 & 0.571 & 0.842\\
    SMN \cite{Wudouban}  & RNN & - & 0.726 & 0.847 & 0.961 &  0.233 & 0.396 & 0.724 & 0.453 & 0.654 & 0.886\\
    DUA \cite{Zhangedc}  & RNN & - & 0.752 & 0.868 & 0.962 &  0.243 & 0.421 & 0.780 & 0.501 & 0.700 & 0.921\\
    DAM \cite{zhou2018multi}  & Trans. & - & 0.767 & 0.874 & 0.969 & 0.254 & 0.410 & 0.757 & 0.526 & 0.727 & 0.933\\
    MRFN \cite{tao2019multi}  & RNN & - & 0.786 & 0.886 & 0.976 & 0.276 & 0.435 & 0.783 & - & - & - \\
    IMN \cite{gu2019interactive}  & RNN & - & 0.794 & 0.889 & 0.974 &  0.262 & 0.452 & 0.789 & 0.621 & 0.797 & 0.964 \\
    IoI \cite{tao2019one}  & RNN & - & 0.796 & 0.894 & 0.974 &  0.269 & 0.451 & 0.786 & 0.563 & 0.768 & 0.950\\
    MSN \cite{yuan2019multi}  & RNN & - & 0.800 & 0.899 & 0.978 & 0.295 & 0.452 & 0.788 & 0.606 & 0.770 & 0.937\\
    G-MSN \cite{lin2020world}  & RNN & \linestack{Grayscale  \\ Data}  & 0.812 & 0.911 & 0.987 &  0.308 & 0.468 & 0.826 & 0.613 & 0.786 & 0.964 \\
    \midrule
    \multicolumn{12}{l}{\textit{PrLM-based methods for fine-tuning}}\\
    BERT \cite{whang2020domain}  & Trans. & - & 0.808 & 0.897 & 0.975 &  0.280 & 0.470 & 0.828 & 0.610 & 0.814 & 0.973\\
    \quad BERT-SS-DA \cite{lu2020improving}  & Trans. & \linestack{Negative \\ Sampling} & 0.813 & 0.901 & 0.977  & 0.280 & 0.491 & 0.843 & 0.648 & 0.843 & 0.980\\ 
    \quad TADAM \cite{xu2021topic}  & Trans. & \linestack{Topic  \\ Segmentation} & 0.821 & 0.906 & 0.978  & 0.282 & 0.472 & 0.828 & 0.660 & 0.834 & 0.975 \\
    \quad PoDS \cite{zhang2021kkt}  & Trans. & Commonsense & 0.828  &  0.912 & 0.981  &  0.287 & 0.468  &  0.845 &  0.633 & 0.810  &   0.967  \\
    ELECTRA \cite{liumdfn}  & Trans. & - & 0.845 & 0.919 & 0.979 & 0.287 & 0.474 & 0.831 & 0.607 & 0.813 & 0.960 \\
    \quad MDFN \cite{liumdfn} &Trans.  & \linestack{Speaker \\ Mask} & 0.866 & {0.932} & {0.984}  & {0.325} & {0.511} & {0.855} & 0.639 &  0.829 & 0.971 \\
    \midrule
    \multicolumn{10}{l}{\textit{Dialogue-related language modeling}}\\
    BERT \cite{whang2020domain}  & Trans. & - & 0.851 & 0.924 & 0.984 & - & - & -  &- & - & - \\
    \quad BERT-VFT \cite{whang2020domain} & Trans. & - & 0.858 & 0.931 & 0.985 & - & - & -  & - & - & -  \\
    \quad SA-BERT \cite{Gusabert}  & Trans. & \linestack{Speaker \\ Embedding} & 0.855 & 0.928 & 0.983 & 0.313 & 0.481 & 0.847 & 0.704 & 0.879 & 0.985\\
    \quad PoDS \cite{zhang2021kkt}  & Trans. & Commonsense & {0.856} & {0.929} & {0.985} &  0.287 &  {0.469} & 0.839 &  {0.671} & {0.842}  &   {0.973}  \\
    \quad DCM \cite{li2021deep}  & Trans. & - & 0.868 & 0.936 & 0.987 &  0.294 & 0.498 & 0.842 & 0.685 & 0.864 & 0.982 \\
    \quad SPIDER \cite{zhang-zhao-2021-structural} & Trans. & - & 0.869 & 0.938 & 0.987 & 0.296 & 0.488 & 0.836 & 0.708 & 0.853 & 0.986 \\
    \quad UMS$_{\text{BERT}}$ \cite{whang2021response}  & Trans. & - & {0.875} & {0.942} & {0.988} & {0.318} & 0.482 & {0.858} & {0.762} & {0.905} & {0.986}  \\
    
    \quad BERT-SL \cite{xu2021learning} & Trans. & - & 0.884 & 0.946 & 0.990 & -  & - & - & 0.776 & 0.919 & 0.991 \\
    ELECTRA  & Trans. & - & 0.861 & 0.932 & 0.985 & 0.301 & 0.499 & 0.836 & 0.673 & 0.835 & 0.974\\
    \quad UMS$_{\text{ELECTRA}}$ \cite{whang2021response}  & Trans. & - & {0.875} & 0.941 & {0.988} & 0.307 & 0.501 & 0.851 & 0.707 & 0.853 & 0.974\\

  \bottomrule
  \end{tabular}}
  \caption{{{Results on Ubuntu, Douban, and E-commerce datasets.}}
  } 
  \label{table:rs_results}
\end{table*}

\section{Empirical Analysis}\label{sec:empirical}
\subsection{Dataset}
We analyze three kinds of dialogue comprehension tasks, 1) response selection: Ubuntu Dialogue Corpus (Ubuntu) \cite{Loweubuntu}, Douban Conversation Corpus (Douban) \cite{Wudouban}, E-commerce Dialogue Corpus (ECD) \cite{Zhangedc}, Multi-Turn Dialogue Reasoning (MuTual) \cite{mutual}\footnote{Because the test set of MuTual is not publicly available, we conducted the comparison with our baselines on the Dev set for convenience.}; 2) conversation-based QA: DREAM \cite{sun2019dream}; 3) conversational machine reading: ShARC \cite{saeidi-etal-2018-interpretation}.\footnote{We only use these typical datasets to save space. Table \ref{sec:dataset_comp} presents a collection of widely-used datasets for the reference of interested readers.}

\subsubsection{Response Selection} 
\paragraph{Ubuntu} consists of English multi-turn conversations about technical support collected from chat logs of the Ubuntu forum. The dataset contains 1 million context-response pairs, 0.5 million for validation, and 0.5 million for testing. In the training set, each context has one positive response generated by humans and one negative response sampled randomly. In the validation and test sets, for each context, there are 9 negative responses and 1 positive response. 
\paragraph{Douban} is different from Ubuntu in the following ways. First, it is an open domain where dialogues are extracted from the Douban Group. Second, response candidates on the test set are collected by using the last turn as the query to retrieve 10 response candidates and labeled by humans. Third, there could be more than one correct response for a context.
\paragraph{ECD} dataset is extracted from conversations between customer and service staff on E-commerce platforms. It contains over 5 types of conversations based on over 20 commodities. There are also 1 million context-response pairs in the training set, 0.5 million in the validation set, and 0.5 million in the test set.

\paragraph{MuTual} consists of 8,860 manually annotated dialogues based on Chinese student English listening comprehension exams.\footnote{MuTual Leaderboard \url{ https://nealcly.github.io/MuTual-leaderboard/}} For each context, there is one positive response and three negative responses. The difference compared to the above three datasets is that only MuTual is reasoning-based. There are more than 6 types of reasoning abilities reflected in MuTual.

\subsubsection{Conversation-based QA}
\paragraph{DREAM} is a dialogue-based multi-choice reading comprehension dataset, which is collected from English exams.\footnote{DREAM Leaderboard \url{ https://dataset.org/dream/}} Each dialogue, as the given context, has multiple questions, and each question has three response options. In total, it contains 6,444 dialogues and 10,197 questions. The most important feature of the dataset is that more than 80\% of the questions are non-extractive, and more than a third of the given questions involve commonsense knowledge. As a result, the dataset is small but quite challenging.

\subsubsection{Conversational Machine Reading}
\paragraph{ShARC} is the current CMR benchmark\footnote{ShARC Leaderboard: \url{ https://sharc-data.github.io/leaderboard.html}}, which contains two subtasks: decision making and question generation{{, as an example shown Table \ref{dialogue_exp} (c). For the first subtask, the machine needs to decide "Yes", "No", "Inquire" and "Irrelevant" given a document consisting of rule conditions}}, initial question, user scenario, and dialog history for each turn. "Yes/No" gives a definite answer to the initial question. "Irrelevant" means that the question cannot be answered with such knowledge base text. If the information provided so far is insufficient for the machine to decide, an "Inquire" decision will be made and we may step into the second subtask and the machine will ask a corresponding question using the under-specified rule span to fill the gap of information. The dataset contains up to 948 dialog trees clawed from government websites. Those dialog trees are then flattened into 32,436 examples. The sizes of train, dev, and test are 21,890, 2,270, and 8,276 respectively.

\begin{table}
{
    \centering\footnotesize
    \setlength{\tabcolsep}{12pt}
    {
        \begin{tabular}{lccc}
            \toprule \textbf{Model} & $\textbf{R}_{4}$@1 & $\textbf{R}_{4}$@2 & \textbf{MRR} \\
            \midrule
            TF-IDF & 0.279 & 0.536 & 0.542 \\
            Dual LSTM  & 0.260 & 0.491 & 0.743 \\
            SMN  & 0.299 & 0.585 & 0.595 \\
            DAM  & 0.241 & 0.465 & 0.518 \\
            \midrule
            GPT-2  & 0.332 & 0.602 & 0.584 \\
            \quad + FT  & 0.392 & 0.670 & 0.629 \\
            BERT & 0.648 & 0.847 & 0.795 \\
            RoBERTa & 0.713 & 0.892 & 0.836 \\
            \quad OCN & 0.867 & 0.958 & 0.926 \\
            ALBERT & 0.847 & 0.962 & 0.916 \\
            \quad GRN-v2 & 0.915 & 0.983 & 0.954 \\
            ELECTRA   &  0.900& 0.979& 0.946 \\
            \quad {MDFN}   & {0.916} & {0.984} & {0.956} \\
            \quad DAPO   & 0.916 & 0.988 & 0.956 \\
             \bottomrule
        \end{tabular}
    }
    \caption{\label{tab:mutual_result} Results on MuTual dataset. The upper and lower blocks present the models w/o and w/ PrLMs, respectively. 
	}
}
\end{table}

\subsection{Evaluation Metrics}
Following \citet{Loweubuntu}, we calculate the proportion of true positive response among the top-$k$ selected responses from the list of $n$ available candidates for one context, denoted as $R_n$@$k$. {{For our tasks, the number of candidate responses $n$ is 10, so we write the metric as $R$@$k$ to save space.}}
For the conversation-based QA task, DREAM, the official metric is accuracy (Acc). Concerning the CMRC task, ShARC evaluates the Micro- and Macro- Acc. for the decision-making subtask. If both the decision is \textsl{Inquire}, BLEU \citep{papineni-etal-2002-bleu} score (particularly BLEU1 and BLEU4) will be evaluated on the follow-up question generation subtask.

\begin{table}
{
    \centering\footnotesize
    \setlength{\tabcolsep}{2.8pt}
    {
        \begin{tabular}{lccc}
            \toprule \textbf{Model} & \textbf{Accuracy} \\
            \midrule
            Stanford Attentive Reader  & 39.8 \\
            Gated-Attention Reader & 41.3 \\
            Word Matching & 42.0 \\
            Sliding Window (SW) & 42.5\\
            Distance-Based Sliding Window & 44.6 \\
            \quad + Dialogue Structure and ConceptNet  & 50.1 \\
            Co-Matching &  45.5 \\
            \midrule
            Finetuned Transformer LM  & 55.5\\
            + Speaker Embedding  & 57.4\\
            EER + FT   & 57.7\\
            BERT-Large + WAE  & 69.0 \\
            RoBERTa-Large + MMM  & 88.9 \\
            ALBERT-xxlarge + DUMA   & 90.4 \\
            \quad + Multi-Task Learning  & 91.8 \\
             \bottomrule
        \end{tabular}
    }
    \caption{\label{tab:dream_result} Results (\%) on DREAM dataset. The upper and lower blocks present the models w/o and w/ PrLMs, respectively.  
	}
}
\end{table}

\begin{table*}
\footnotesize
\centering\centering\setlength{\tabcolsep}{2.1pt}
\begin{tabular}{lcccccccc}
\toprule
\multirow{3}{*}{Model} &
\multicolumn{4}{c}{Dev Set} & \multicolumn{4}{c}{Test Set}\\
&\multicolumn{2}{c}{Decision Making} & \multicolumn{2}{c}{Question Gen.} & \multicolumn{2}{c}{Decision Making} & \multicolumn{2}{c}{Question Gen.}\\
\cmidrule{2-5}
\cmidrule{6-9}
 & Micro & Macro & BLEU1 & BLEU4 & Micro & Macro & BLEU1 & BLEU4 \\ 
\midrule
NMT \citep{saeidi-etal-2018-interpretation}  &-&-&-&-& 44.8 & 42.8 & 34.0 & 7.8\\
CM \citep{saeidi-etal-2018-interpretation}&-&-&-&-  & 61.9 & 68.9 & 54.4 & 34.4\\
UcraNet \citep{verma-etal-2020-neural} &-&-&-&- & 65.1&71.2&60.5&46.1\\
\midrule
BERTQA \citep{zhong-zettlemoyer-2019-e3}   &68.6&73.7&47.4&54.0  & 63.6   & 70.8   & 46.2 & 36.3 \\
BiSon  \citep{lawrence-etal-2019-attending} &66.0&70.8&46.6&54.1&66.9&71.6&58.8&44.3 \\
E$^3$ \citep{zhong-zettlemoyer-2019-e3}  &68.0&73.4&67.1&53.7 & 67.7  & 73.3    & 54.1 & 38.7  \\
EMT \citep{gao-etal-2020-explicit}&73.2&78.3&67.5&53.2 &69.1 &74.6&63.9& 49.5 \\
DISCERN \citep{gao-etal-2020-discern} &74.9&79.8&65.7&52.4&73.2&78.3 & 64.0 &49.1 \\
DGM \citep{ouyang2020dialogue}& 78.6 & 82.2 & 71.8 & 60.2 & 77.4 & 81.2 & 63.3 & 48.4\\
\bottomrule
\end{tabular}
\caption{Results on the the dev set and blind held-out test set of ShARC tasks for decision making and question generation. Micro and Macro stand for Micro Accuracy and Macro Accuracy. The upper and lower blocks present the models w/o and w/ PrLMs, respectively.} \label{table:e2e}
\end{table*}

\subsection{Observations}
Tables \ref{table:rs_results}-\ref{table:e2e} present the benchmark results on six typical dialogue comprehension tasks, including three response selection tasks, Ubuntu \cite{Loweubuntu}, Douban \cite{Wudouban}, ECD \cite{Zhangedc}, Mutual \cite{mutual}, one conversation-based QA task, DREAM \cite{sun2019dream}, and one conversation machine reading task consisting of decsion making and question generation, ShARC \cite{saeidi-etal-2018-interpretation},  from which we summarize the following observations:\footnote{The evaluation results are collected from published literature \cite{zhang2021kkt,whang2021response,xu2021learning,lin2020world,Loweubuntu,liu2020graph,liumdfn,Lidapo,sun2019dream,wan2020multi}.}

1) \textbf{{{Interaction methods generally yield better performance than single-turn models.}}} In the early stage without PrLMs, separate interaction (Framework 2) commonly achieves better performance than the simple concatenated matching (Framework 1), verifying the effectiveness of attention-based pairwise matching. However, multi-turn matching networks (separate interaction) perform worse than PrLMs-based ones (Framework 3), illustrating the power of contextualized representations in context-sensitive dialogue modeling.

2) {{\textbf{Dialogue-related pre-training helps make PrLM better suitable for dialogue comprehension.}}} Compared with general-purpose PrLMs, dialogue-aware pre-training (e.g., BERT-VFT \cite{whang2020domain}, SA-BERT \cite{Gusabert}, PoDS \cite{zhang2021kkt}) can further improve the results by a large margin. In addition, task-oriented pre-training (e.g., DCM \cite{li2021deep}, UMS \cite{whang2021response}, BERT-SL \cite{xu2021learning}) even shows superiority among the pre-training techniques.

3) {{\textbf{Discriminative modeling beats generative methods.}}} Empirically, for the concerned dialogue comprehension tasks, retrieval-based or discriminative methods commonly show better performance than generative models such as GPT \cite{radford2018improving}.

4) {{\textbf{Data augmentation from negative sampling has attracted interests to enlarge corpus size and improve response quality.}}} Among the models, G-MSN \cite{lin2020world}, BERT-SS-DA \cite{lu2020improving}, ELECTRA+DAPO \cite{Lidapo} show that training/pre-training data construction, especially negative sampling is a critical influence factor to the model performance. 

5) {{\textbf{Context Disentanglment helps discover the essential dialogue structure.}}} SA-BERT \cite{Gusabert} and MDFN \cite{liumdfn} indicate that modeling the speaker information is effective for dialogue modeling. Further, EMT \cite{gao-etal-2020-explicit}, Discern \cite{gao-etal-2020-discern}, and DGM \cite{ouyang2020dialogue} indicate that decoupling the dialogue context into elementary discourse units (EDUs) and model the graph-like relationships between EDUs would be effective to capture inner discourse structures of complex dialogues.

6) {{\textbf{Extra knowledge injection further improves dialogue modeling.} From Table \ref{tab:dream_result}, we see that external knowledge like commonsense would be beneficial for dialogue modeling of dialogue systems \cite{Lidapo}. Recent studies also show that interactive open-retrieval of relevant knowledge is useful for reducing the hallucination in conversation \cite{shuster2021retrieval}.}}



\section{{{Frontiers of Training Dialogue Comprehension Models}}}
\subsection{Dialogue Disentanglment Learning} 
Recent widely-used PrLM-based models deal with the whole dialogue,\footnote{Because PrLMs are interaction-based methods, thus encoding the context as a whole achieves better performance than encoding the utterances individually.} which results in entangled information that originally belongs to different parts and is not optimal for dialogue modeling, especially for multi-party dialogues \cite{yang2019friendsqa,li2020transformers,li2020molweni}. Sequence decoupling is a strategy to tackle this problem by explicitly separating the context into different parts and further constructing the relationships between those parts to yield better fine-grained representations.
One possible solution is splitting the context into several topic blocks \cite{xu2021topic,lu2020improving}. However, there existing topic crossing, which would hinder the segmentation effect. Another scheme is to employ a masking mechanism inside self-attention network \cite{liumdfn}, to limit the focus of each word only on the related ones, such as those from the same utterance, or the same speaker, to model the local dependencies, in complement with the global contextualized representation from PrLMs. Further, recent studies show that explicitly modeling discourse structures and action triples \cite{gao2020discern,ouyang2020dialogue,chen2021structure,li2021dadgraph,feng2020dialogue} would be effective for improving dialogue comprehension. 




\subsection{Dialogue-aware Language Modeling}
Recent studies have indicated that dialogue-related language modeling can enhance dialogue comprehension substantially \cite{Gusabert,li2021deep,zhang2021kkt,whang2021response,xu2021learning}. However, these methods rely on the dialogue-style corpus for the pre-training, which is not always available in general application scenarios. Given the massive free-form and domain-free data from the internet, how to simulate the conversation, e.g., in an adversarial way, with the general-purpose and general-domain data, is a promising research direction. Besides transferring from general-purpose to dialogue-aware modeling, multi-domain adaption is another important topic that is effective to reduce the annotation cost and achieve robust and scalable dialogue systems \cite{qin2020dynamic}.
\subsection{High-quality Negative Sampling}
Most prior works train the dialogue comprehension models with training data constructed by a simple heuristic. They treated human-written responses as positive examples and randomly sampled responses from other dialogue contexts as equally bad negative examples, i.e., inappropriate responses \cite{Loweubuntu,Wudouban,Zhangedc}. As discussed in Section \ref{sec:empirical}, data construction is also critical to the model capacity. The randomly sampled negative responses are often too trivial, making the model unable to handle strong distractors for dialogue comprehension. Intending to train a more effective and reliable model, there is an emerging interest in mining better training data \cite{lin2020world,Lidapo,su2020dialogue}.

\section{{{Open Challenges}}}
{{Though there are extensive efforts that have been made, with impressive results obtained on many benchmarks for dialogue comprehension, there are still various open challenges.}}

\paragraph{{{Temporal Reasoning}}}
{{Daily dialogues are rich in events, which in turn requires understanding temporal commonsense concepts interwoven with those events, such as duration, frequency, and order. There are preliminary attempts to investigate temporal features like utterance order and topic flow. However, such features are too shallow to reveal the reasoning chain of events. \citet{qin-etal-2021-timedial} showed that the best dominant variant of PrLM like T5-large with in-domain training still struggles on temporal reasoning in dialogue which relies on superficial cues based on existing temporal patterns in context.}}

\paragraph{Logic Consistency}
Logic is vital for dialogue systems which not only guarantees the consistency and meaningfulness of responses but also strengthens the mode with logical reasoning abilities. Existing interaction-based commonly focus on capturing the semantic relevance between the dialogue context and the response but usually neglect the
logical consistency during the dialogue that is a critical issue reflected in dialogue models \cite{mutual}. The widely-used backbone PrLM models are trained from plain texts with simple LM objectives, which has shown to suffer from adversarial attacks \cite{liu2020robust} easily and lack the specific requirements for dialogue systems such as logic reasoning. 

\paragraph{Large-scale Open-retrieval}
The current mainstream dialogue tasks often assume that the dialogue context or background information is provided for the user query. In real-world scenarios, a system would be required to retrieve various types of relevant information such as similar conversation history from a large corpus or necessary supporting evidence from a knowledge base to respond to queries interactively. Therefore, how to retrieve accurate, consistent, and semantically meaningful evidence is critical. Compared with the open-domain QA tasks, open-retrieval dialogues raise new challenges of both efficiency and effectiveness mostly due to the human-machine interaction features. 


\paragraph{{{Dialogue for Social Good}}}
{{To effectively solve dialogue tasks, we are often equipped with specific and accurate information, from which the datasets are often crawled from real-world dialogue histories that require many human efforts. Domain transfer is a long-standing problem for the practical utility of dialogue systems that is far from being solved. As reflected in Table \ref{sec:dataset_comp}, dialogue corpora are often restricted in specific domains. There are many domains that have not been paid litter attention to due to the lack of commercial interests or lack of annotated data. Besides, few attention has been paid to low-resource communities, and high-quality dialogue corpus is generally scarce beyond English communities.}}


\bibliography{tacl2018}
\bibliographystyle{acl_natbib}

\end{document}